\documentclass[wcp]{jmlr}

\usepackage{longtable}

\usepackage{booktabs}

\graphicspath{{imgs/}}
\DeclareGraphicsExtensions{.pdf,.jpeg,.png}
\usepackage{textcomp}
\def\BibTeX{{\rm B\kern-.05em{\sc i\kern-.025em b}\kern-.08em
    T\kern-.1667em\lower.7ex\hbox{E}\kern-.125emX}}
\usepackage{microtype}
\usepackage{graphicx}
\usepackage{booktabs} 
\usepackage{amsmath}
\usepackage{amssymb}
\usepackage{mathrsfs}  
\usepackage{hyperref}

\usepackage{amsfonts}
\usepackage{booktabs}
\usepackage{siunitx}
\usepackage[switch]{lineno}   
\usepackage{amsmath}  
\usepackage{etoolbox} 
\newcommand*\linenomathpatch[1]{%
  \cspreto{#1}{\linenomath}%
  \cspreto{#1*}{\linenomath}%
  \csappto{end#1}{\endlinenomath}%
  \csappto{end#1*}{\endlinenomath}%
}
\linenomathpatch{equation}
\linenomathpatch{gather}
\linenomathpatch{multline}
\linenomathpatch{align}
\linenomathpatch{alignat}
\linenomathpatch{flalign}
\usepackage{enumitem}
\usepackage{indentfirst}
\usepackage{lineno}

\makeatletter
\let\Ginclude@graphics\@org@Ginclude@graphics 
\makeatother

\title{Reinforcement Learning for Solving Stochastic Vehicle Routing Problem}

 \author{\Name{Zangir Iklassov} \Email{zangir.iklassov@mbzuai.ac.ae}\\
  \Name{Ikboljon Sobirov} \Email{ikboljon.sobirov@mbzuai.ac.ae}\\
  \Name{Ruben Solozabal} \Email{ruben.solozabal@mbzuai.ac.ae}\\
  \Name{Martin Tak\'a\v{c}} \Email{martin.takac@mbzuai.ac.ae}\\
  \addr MBZUAI, UAE, Abu-Dhabi}

\begin{document}

\setlength{\abovedisplayskip}{3pt}
\setlength{\belowdisplayskip}{3pt}

\maketitle

\begin{abstract}
This study addresses a gap in the utilization of Reinforcement Learning (RL) and Machine Learning (ML) techniques in solving the Stochastic Vehicle Routing Problem (SVRP) that involves the challenging task of optimizing vehicle routes under uncertain conditions. We propose a novel end-to-end framework that comprehensively addresses the key sources of stochasticity in SVRP and utilizes an RL agent with a simple yet effective architecture and a tailored training method. Through comparative analysis, our proposed model demonstrates superior performance compared to a widely adopted state-of-the-art metaheuristic, achieving a significant 3.43\% reduction in travel costs. Furthermore, the model exhibits robustness across diverse SVRP settings, highlighting its adaptability and ability to learn optimal routing strategies in varying environments. The publicly available implementation of our framework serves as a valuable resource for future research endeavors aimed at advancing RL-based solutions for SVRP. 
\end{abstract}

\section{Introduction}
\label{introduction}

Reinforcement Learning (RL) studies machine learning (ML) problems with a target to train algorithms to make decisions with a reward maximization function through interactions with their environment. With notable success of RL in numerous applications such as gaming \cite{silver2017mastering} and robotics \cite{Andrychowicz2020LearningDI}, their use cases in combinatorial optimization problems are gaining great attention. Combinatorial optimization problems are one of the most challenging optimization problems that is traditionally addressed with heuristics or solvers (e.g. Google OR tools). Heuristic approaches are generally fast but they are only locally optimal; solvers can provide globally optimal solutions but are significantly slow. RL approaches can resolve both of these issues with computationally efficient models that do not require human intervention to learn black-box heuristics. Such models, however, should be carefully formulated such that the environment and training settings are well-established for the problem at hand. 

The vehicle routing problem (VRP) is a classic combinatorial optimization problem which aims to find an optimum route for a set of vehicles with limited capacity that deliver goods or services to a group of customers such that the overall distance traveled is minimized. \cite{nazari2018reinforcement} proposed the first RL model that can tackle the VRP using a pointer network.

The \textbf{stochastic vehicle routing problem} (SVPR) is a subtype of VPR problems, and unlike deterministic VRP, it is even more challenging on account of uncertain parameters, which can only be determined after the vehicle completes several routes. While the SVRP poses certain difficulties, such a formulation of the VRP is more representative to the real-world applications that are commonly encountered in industry \cite{inbook, GENDREAU19963}. Several classical approaches to solving the SVRP have been studied in~\cite{Laporte1993TheIL, Pichpibul2013AHA, Goel2019VehicleRP}; however, the problem has not been yet investigated using deep reinforcement learning.

A particular parameters of the SVRP, known as correlated stochastic demands and travel times, make the problem distinctly challenging due to the randomly variable customer demands that are non-independent/correlated based on certain underlying factors \cite{Bahaabadi2021}. Solving the SVRP with correlated stochastic demands and travel times offers numerous benefits to companies that operate in transportation and logistics sectors, including reduction in carbon emissions and costs. In this work, our goal is to leverage RL to tackle the SVRP with correlated stochastic demands and travel times such that the formulation is closely aligned with the real-world settings.

In this research endeavor, we aim to enhance the applicability of our framework \cite{iklassovSST23} to real-world supply-chain scenarios by incorporating two important sources of stochasticity, namely stochastic demand and stochastic travel costs. We conduct experiments to assess the dependence of our proposed model on different settings of SVRP environment, which are frequently explored in classical research. It is important to note that the proposed architecture is designed to be flexible and capable of addressing SVRP instances with an arbitrary number of vehicles. Our specific {\bf contributions} to this field are summarized as follows:
\begin{itemize}
    \item In this study, we address the Stochastic Vehicle Routing Problem and propose a \textbf{an RL-based approach} to solve it. Our model incorporates stochastic customer demands and travel costs, which are crucial factors in real-world logistics scenarios. We compared our RL model to state-of-the-art approaches and demonstrated its superiority, achieving a \textbf{3.43\% reduction in travel cost} compared to the best-performing classical model  \cite{Goel2019VehicleRP}.
    \item We conducted comprehensive experiments to evaluate the performance of our model across various VRP settings. We observed that incorporating weather information, which introduced correlations among customers, enabled our RL agent to achieve up to a \textbf{2\% improvement in routing strategies}.
    \item Overall, our research showcases the effectiveness of RL in addressing the SVRP and its potential to outperform traditional approaches used in industry. The results highlight the \textbf{significant cost reductions and potential environmental benefits} that can be achieved by adopting RL-based routing solutions using \textbf{shared open-source framework}\footnote{\url{https://github.com/Zangir/SVRP}}.
\end{itemize}

\section{Related Work}
\label{related_work}

Using RL methods for combinatorial optimization problems has been investigated in numerous studies. \cite{Bello2017NeuralCO} pioneered the use of RL to address the Traveling Talesman (TSP) and Knapsack problems, reporting a performance exceeding typical heuristics. In fact, their approach reached the results close to global optimum. \cite{Oroojlooyjadid2022ADQ} investigated the Beer Game problem using a Q-learning model, which can essentially be extended to solve decentralized multi-agent combinatorial problems. \cite{Tricks2018AND} also dives into the application of RL techniques to search for optimal solutions for combinatorial problems such as Knapsack, Secretary, and Adwords. Such studies encourage the potential of RL applications in the field of combinatorial optimization problems. 

Deterministic VRP is no exception to benefit from RL techniques in recent years. \cite{nazari2018reinforcement} reported the first work that leverage an RL method to solve VRP, authors proposed pointer network based architecture to solve deterministic VRP that gives superior performance compared to baseline heuristic and metaheuristic methods. \cite{Lu2020ALI} developed a metaheuristic RL-based agent that supervises simple rules in the search for VRP solution. The approach sets the state-of-the-art (SOTA) with the lowest cost for VRP instances with 50-100 customers; however, it requires a domain knowledge from the researcher to generate a set of plausible heuristics. To alleviate all this problem, \cite{Wang2021AlphaTLT} proposed an RL framework that reaches a new SOTA performance for larger scale instances of VRP at 500, 1000, and 2000 customers.

While there are numerous research papers on the deterministic VRP that leverage RL methods, the use of such approaches in SVRP problems still remain a challenge. The current solutions for SVPR, as mentioned in~\cite{Toth2014VehicleRP}, rely heavily on the classical algorithms for SVPR, which can be summarized in the following categories:
\begin{itemize}
  \item \textbf{Branch-and-Bound} algorithms - Branch-and-Cut \cite{Gauvin2014ABA}, Branch-and-Price \cite{Christiansen2007ABA}, Integer L-shaped \cite{Laporte1993TheIL}.
  \item \textbf{Heuristics} - LKH3 \cite{Helsgaun2017AnEO}, Clarke-Wright \cite{Pichpibul2013AHA}.
  \item \textbf{Metaheuristics} - Tabu Search \cite{Li2020AnIT}, Ant-Colony Optimization \cite{Goel2019VehicleRP}.
\end{itemize}

Compared to the other branch-and-branch family members, the integrated L-shaped algorithm \cite{Laporte1993TheIL} proved to be superior for the SVRP problem. The cost is substituted with a $\theta$ variable that acts as a lower bound evaluator through the application of branch-and-branch method. While this is desirable, its performance is computation-heavy with regards to the number of vehicles, making it undesirable in real-life applications. Among the many solutions for the VRP, LKH3 \cite{Helsgaun2017AnEO} yields unmatchable performance for the deterministic VRP; however, its performance becomes unstable for SVRP. Clarke-Wright algorithm \cite{Pichpibul2013AHA} addresses that issue, performing similarly well for both deterministic and stochastic VRP, while being a simple heuristic-based approach. The Clarke-Wright algorithm has become the go-to reference/baseline model for comparison in the SVRP literature due to its consistency and longevity.  Metaheuristic methods slowly emerged to become prevalent in the literature as well, with the Ant-Colony optimization problem and such methods' SOTA performance on the problem \cite{Goel2019VehicleRP}.

A separate subfield within the literature on SVRP involves instances with correlated stochastic parameters. In order to ensure the efficacy of classical methods for SVRP instances with non-i.i.d. parameters, it is crucial to consider the correlation in the design of the algorithm. For instance, the pheromone update formula in ACO 
should be adjusted to reflect the correlation between parameters \cite{Bahaabadi2021}. Similarly, Branch-and-Bound algorithms must be modified to handle this correlation \cite{Bomboi2021OnTS}. In general, classical algorithms have been shown to generally exhibit inferior performance on this type of problem due to the assumption of i.i.d. parameters in their design. 

In recent years, a limited number of studies have applied ML algorithms to the SVRP. In \cite{Secomandi2000ComparingNP}, a policy-based model was proposed to address the SVRP with stochastic demands, where the authors trained a simple linear model and achieved improved performance compared to heuristic methods. In \cite{Niu2021AnIL}, the authors utilized decision trees to solve the SVRP, while \cite{Niu2022MultiobjectiveEA} employed an evolutionary algorithm based on a radial basis network. Furthermore, \cite{Joe2020DeepRL} was the first study to use deep reinforcement learning to address the SVRP with stochastic customers. \cite{Mustakhov2023DeepRL} extended this by considering stochastic customers and aiming to maximize the total goods served within a defined time frame. Notably, \cite{iklassovSST23} were the first to employ Reinforcement Learning (RL) for solving SVRP with stochastic correlated demands, achieving state-of-the-art performance in the considered problem domain. In our pursuit, we endeavor to augment this approach by integrating stochastic travel times, thereby subjecting the model to comprehensive evaluation.

\section{Background}
\label{background}

One of the most widely observed NP-hard combinatorial problems is the \textbf{Vehicle Routing Problem} (VRP). The VRP solves the optimization problem of delivering goods from a depot to customers with known locations and demands. The goal of the method is to minimize the overall travel cost required to meet the demands of all the customers. Several studies delved into the VRP optimization using different methods and heuristics \cite{Laporte2009FiftyYO}; but the initial VRP formulation was proposed by \cite{Dantzig1959TheTD}.


The \textbf{Stochastic Vehicle Routing Problem} (SVRP) is a more challenging task due to the uncertainty in the parameters of the method, which can only be known after the completion of a route. Unlike the VRP, the SVRP is stochastic in such parameters, making it a harder task; however, because of its greater applicability in real-world industrial problems, it attracts more community in industry and research. It can be a better representative model of the real-life routing settings, and therefore, there are several works investigating the problem \cite{inbook, GENDREAU19963}. \cite{Toth2014VehicleRP} attributes three main sources as the stochasticity of the SVRP:
\begin{enumerate}[noitemsep]
  \item \textbf{stochastic demands} (VRPSD), i.e., uncertain customer demands,
  \item \textbf{stochastic customers} (VRPSC), i.e., uncertain presence/absence of customers,
  \item \textbf{stochastic travel times} (VRPSTT), i.e., uncertain travel costs for various routes. 
\end{enumerate}


The deterministic nature of VRP cannot suffice as an optimum solution for SVRP, claimed \cite{Louveaux1998AnIT}, and proposed a formulation that takes the stochasticity of the problem into account using a mathematical representation.


\subsection{Classical Formulation}

The objective of the problem is to
\begin{align}
\text{minimize}\qquad &\textstyle{\sum}_{i, j \in C} c_{i j} x_{i j}+\mathscr{R}(x). \label{eq:1}
\end{align}

The problem at hand is characterized by a set of notations as follows: $N$ - set of customers and depot, $C$ - customers set, $c_{i j}$ - stochastic travel cost between nodes $i$ and $j$, $c_{ijs}$ - $s^{th}$ realization of $\{ij\}$ travel cost, $\xi_{i}$ - stochastic demand of customer $i$, $\mu_{is}$ - $s^{th}$ realization of the demand of customer $i$, $K$ - set of vehicles, $Q$ - maximum capacity of vehicle, $x_{i j}$ - binary variable that shows whether $(i, j)$ is used in the route.

In this formulation, the SVRP is modeled on a complete, undirected graph $G=(N, X)$. A depot node (i.e. node 0) and customer nodes are represented as the set of nodes $N=\{0,1, \ldots, n\}$. The depot functions as the base for the set of $K$ vehicles with an initial loading value of $Q$. A stochastic demand variable $\xi_{i}$, including its $s^{th}$ realization characterized as $\mu_{is}$, exists per customer node $i \in C=N \backslash{0}$. Arcs, given as $X={(i, j): i, j \in N, i<j}$, denote the associations between nodes; the travel cost of an arc $(i, j) \in X$ in a route is $c_{ij}$. The representation $x_{i j}$ is an action of an agent with $x_{i j}$=1 when an arc $(i, j)$ is traveled in the route and 0 otherwise. In case of situations when a vehicle fails to fulfill the demand of a customer due to a shortage of load, the recourse cost $\mathscr{R}(x)$ accounts for it. To be precise, the term is responsible for the cost incurred for traversing to the depot and back to refill in order to meet the customer demand. The detailed description of the classical problem, including all the constraints and additional information, can be found in the supplementary material.


\subsection{Baselines}

Several methods existent in the literature serve as baselines for the SVRP. The \textbf{Clarke-Wright} (CW) heuristic \cite{Pichpibul2013AHA} is such an approach that is based on the measure, so-called \textit{savings}, that estimates the decrease in the total travel cost that occurs due to the merge of two customer nodes into a single route. The \textit{saving} amount for two customer nodes $i$ and $j$ is computed as $saving_{ij}=\mathbb{E} [c_{0i}]+\mathbb{E} [c_{j0}]-\mathbb{E} [c_{ij}], $ where $c_{0i}$ and $c_{0j}$ are the traversing costs between depot and customer $i$ and $j$, respectively, and $c_{ij}$ is the traversing cost between them. These values are then collected for all the customer pairs in a savings list, sorted in decreasing order. The technique for merging the routes starts with picking the highest-value pair from the list. Provided that the expected demand given as $\mathbb{E} [\xi_{i}] + \mathbb{E} [\xi_{j}]$ is within the vehicle capacity $Q$, the two customer nodes are merged together in a single route. This process repeats until no viable merging is left in the savings list.


Another well-established optimization approach is \textbf{Tabu Search} (TS), which uses metaheuristics to solve the SVRP. With this approach, a random feasible solution is generated randomly, and its total cost is calculated. Using a set of predefined neighborhood heuristic procedures on the proposed solution, a new list of candidate solutions is generated over a certain number of iterations defined as $k_{Tabu, max}$. The algorithm then outputs the best candidate from the set of visited feasible solutions.


In~\cite{Li2020AnIT}, the authors demonstrate that choosing the right set of four heuristics (Specific neighborhood, 2-opt, Swap-operation, Reallocate-operation) can outperform other sets of heuristics studied in the older literature.  


\textbf{Ant Colony Optimization} (ACO) is also a metaheuristics baseline that mimics how ants in nature behave foraging for food. Solution agents act as ants traveling the search space by investigating and exploiting the possible routes. When an agent finds a feasible solution, it leaves a pheromone on the arcs that build this route. The probability of an arc being a better candidate depends on the concentration of the pheromone laid by the agent, which can later be followed by subsequent agents. In other words, the higher the concentration of the pheromone, the better quality that arc has as a candidate. The pheromone fades away over time, again imitating the natural ant behavior. The search for new solutions goes on as ants explore more arcs to follow probabilistically.

\section{Method}

{\bf Objective.} We propose a problem formulation where the objective is to 
\begin{align}
\text{minimize} \quad &\tfrac{1}{|S|}\textstyle{\sum}_{s \in S} (\textstyle{\sum}_{i, j \in C} c_{ijs} x_{ij}+2\textstyle{\sum}_{i \in C} r_{i}c_{0is}), \nonumber 
\end{align}
where the agent minimizes the travel cost in the context of a set of scenarios, denoted as $S$. The agent aims to solve this problem by making decisions on action variables, specifically $x_{ij}$ and $r_i$. Here, $r_i$ represents a binary recourse action that determines whether the vehicle needs to return to the depot after visiting customer $i$ in order to replenish its capacity and prevent potential issues with future customers along the route.

{\bf Stochastic Variables.} The agent lacks prior knowledge of the distribution of stochastic variables and does not have access to the specific realization $s$ of the demand $\mu_{is}$ or the travel cost $c_{ijs}$. The agent can only observe the value of $\mu_{is}$ after visiting customer $i$ and the value of $c_{ijs}$ after traversing the arc $\{ij\}$. However, the agent has the ability to estimate the values of stochastic variables by leveraging the knowledge of the observable variables. We define a set of random variables $W$ that influence the realization of demand and travel costs in the following manner:
\begin{align}
\theta_{i} = \bar{\theta}_{i} + \textstyle{\sum}_{m} \textstyle{\sum}_{n} \alpha_{imn} w_{m} w_{n} +\epsilon_i \text{,   }\ \theta_{i} \in \{ \xi_{i}, c_{i0}, ..., c_{iN}\},
\nonumber
\end{align}
where we define $\theta_i$ as a stochastic variable representing the demand of customer $i$ or the travel cost from customer $i$ to other nodes. The value of $\theta_i$ is determined by a fixed term $\bar{\theta}_i$, which is augmented by the weighted interaction between variables $w_m \in W$ and $w_n \in W$, as well as random noise $\epsilon_i$. It is important to note that the agent does not possess knowledge of the distribution of the variables in $W$ or their specific influence on $\theta_i$. However, the agent does have access to the realizations of $W$ variables, allowing him to implicitly learn the mapping from $W$ variables to stochastic variables to enhance the routing strategy.

{\bf Weather Variables.} This study focuses on a set of three $W$ variables referred to as weather variables, namely temperature, pressure, and humidity. This selection is motivated by real-world scenarios where the agent lacks prior knowledge of demand and travel costs but can estimate them by observing the prevailing weather conditions, which exert a direct influence on these factors. For instance, the demand for ice-cream and the travel time between ice-cream shops may vary based on the temperature. By incorporating these weather variables, we aim to create a realistic environment where the agent utilizes observable weather information to estimate the associated stochastic variables effectively. It is important to acknowledge that the presented formulation provides an abstract representation of the problem, and in practice, there may exist additional stochastic variables beyond demand and travel time that impact the agent's costs. Moreover, there is the possibility of incorporating additional $W$ variables that the agent assumes to have an influence on the environment. 

\subsection{Routing Policy}

\textbf{State}. The proposed SVRP formulation defines the environment as a state $(I^t_s, h^t_k)$ associated with a particular realization $s$ and vehicle $k$,
\begin{align}
 I^t_s \doteq \{(W, d_{i}^t, c_{i0s}, ..., c_{iNs}), i \in N\} \text{,   }\ h^t_k \doteq (q^t_k, p^t_k), k \in K. \nonumber
\end{align}
The state is dynamic and evolves over time $t \in \{1, ..., T\}$, where $T$ represents the time step when all vehicles have returned to the depot and all customer demands have been fulfilled. $I^t_s$ is characterized by a set of weather variables $W$, the current demand of customer $i$ at time $t$, denoted as $d_{i}^t$, and the travel cost realizations $c_{ijs}$ between all nodes $\{ij\}$. The weather variables and travel costs remain constant throughout the time horizon, while the demand $d_i^t$ takes on the value of the demand realization $\mu_{is}$ at $t = 1$, and $0$ at $t = T$ for all customers $i \in N$. $h^t_k$ is characterized the current load $q^t_k$ and the position $p^t_k$ of each vehicle $k \in K$ at time $t$.

{\bf Action.} At each time step $t$, the agent selects an action $a^{t}_k$ for each vehicle, where the action represents the choice of the next node $i \in N$ to visit. The decision-making process is dynamic and governed by a policy $\pi$, such that $a^{t}_k \sim \pi(\cdot| I^t_s, h^t_k)$. In other words, the policy determines the probability distribution of selecting the next node for each vehicle based on the current state ($I^t_s, h^t_k)$.

{\bf Transitions.} Upon executing the action $a^t_k$, the positions of the vehicles are updated to $p^{t+1}_k$ based on the nodes they move to. The representation of vehicle positions can be encoded using various techniques, such as coordinates on a 2D map or one-hot vectors. The demands and loads for vehicle $k$ that visits the customer $i$ are updated according to the following formulas:
\begin{align}
d^{t+1}_i = \max\left\{0,d^{t}_i-q^t_k\right\} \text{, }\
d^{t+1}_j = d^{t}_j, \text{ for }j\neq i, \text{ and }\ 
q^{t+1}_k = \max\left\{0,q^t_k-d^{t}_i\right\}. \label{eq:vrp:dem-load1}
\end{align}

\textbf{Objective.} The agent aims to derive an optimal policy $\pi$ with parameters $\Theta$ by employing the Reinforce algorithm, which seeks to minimize the cumulative expected cost
\begin{align}
    \mathcal{J}^{\pi}(\Theta) = \mathbb{E} [\textstyle{\sum}_t^T C_t(I^t_s, a^t_k)]. \nonumber
\end{align}
The cost function $C_t(I^t_s, a^t_k)$ is defined as the sum of traversal costs for each vehicle $k$ moving from node $i_{k}$ to node $j_{k}$ after executing action $a^t_{k}$. It is calculated as $\textstyle{\sum}_{k=1}^{K} c_{i_{k}j_{k}s}$. In cases where vehicle $k$ requires recourse action due to a failure, an additional recourse cost of $2c_{0j_{k}s}$ is incurred. The objective function is accumulated over a set of scenarios $S$, each consisting of $T_s$ time steps, and the gradient of the objective function is computed as follows:
\begin{align}
\nabla_\theta{\hat{\mathcal{J}}^{\pi}}(\Theta) \approx \nonumber \tfrac{1}{S} \textstyle{\sum}_{s=1}^{S} \textstyle{\sum}_{t=1}^{T_s} ( (C(I^t_s, h^t_k) - b_\phi(I^t_s, h^t_k)) \cdot  \nabla_\Theta \log{\pi_\Theta(a^t_k | I^t_s, h^t_k)}), \nonumber
\end{align}
where ${b_\phi}(I^t_s, h^t_k)$ denotes the baseline function with parameters $\phi$ that are trained to minimize 
$
    L(\phi) =  \tfrac{1}{S} \textstyle{\sum}_{s}^{S}  \textstyle{\sum}_{t=1}^{T_s} ||b_\phi(I^t_s, h^t_k) - C(I^t_s, h^t_k) ||^2. $
\subsection{Environment Settings}

The model's robustness and performance improvement over time need to be evaluated across various environmental settings that closely resemble real-life scenarios. This experimentation is crucial to assess its ability to adapt and excel in diverse conditions. Therefore, we aim to examine key environmental factors that hold significance in classical SVRP research.

{\bf Inference.} The performance evaluation of the model can be conducted by investigating its reliance on different inference strategies, encompassing widely employed methods like greedy sampling, random sampling, and beam search. In the case of greedy sampling, the action with the highest probability is chosen at each time step $t$ to construct the routing solution. Random sampling generates multiple solutions by randomly selecting $n_s$ actions in accordance with their respective probability distributions at each time step. In beam search, the top $n_b$ cumulative most probable actions are selected. The utilization of these distinct inference strategies may yield disparate routing solutions, thereby leading to diverse performance outcomes. 

{\bf Signal Ratio.} The three components of $\theta_i$, namely the constant term, weather-related factors, and random noise, are likely to exert varying influences on the distribution of stochastic variables. In order to quantify this influence, we introduce signal ratio variables $A_{i}$, $B_{i}$, $\Gamma_{i}$ and investigate the impact of different signal ratio values on the model's performance.
\begin{align*}
A_{i} &= \tfrac{\bar{\theta_{i}^{2}}}{T_i} &
B_{i} &= \tfrac{\mathbb{E}\left[(\sum_{m}\sum_{n}\alpha_{imn}w_{m}w_{n})^{2}\right]}{T_i},  &
\Gamma_{i} &= \tfrac{\mathbb{E}\left[\epsilon_{i}^{2}\right]}{T_i}, 
\end{align*}
\begin{align}
T_{i} &= \bar{\theta_{i}^{2}} + \mathbb{E} [(\textstyle{\sum}_{m}\textstyle{\sum}_{n}\alpha_{imn}w_{m}w_{n})^{2} ] + \mathbb{E}\ [\epsilon_{i}^{2} ]. \nonumber
\end{align}

{\bf Fill rate.} The performance of the model may be influenced by the maximum capacity of each vehicle. To quantify its impact, we introduce a fill rate variable, denoted as $\Phi = \frac{Q}{\mathbb{E}[\xi_i]}$, which represents the relative value of the maximum capacity in relation to the expected value of the total demand.

{\bf A Priori vs Reoptimization.} In SVRP, there are two distinct approaches that are both significant in industry and warrant consideration. The first approach, known as a priori, involves the agent constructing the entire route without prior knowledge of the realizations of stochastic variables. On the other hand, the reoptimization approach allows the agent to unveil the realizations of variables as it visits the corresponding customers and arcs. This enables the agent to dynamically update the routes based on the acquired information, thus addressing the problem in an online fashion.

{\bf Variable Estimates.} Prior to constructing the route, the agent can utilize estimates of the stochastic variables as inputs, denoted as $\hat{\theta_i}$. In this study, we propose two types of estimates: the constant estimate and the k-NN estimate. The constant estimate simply represents the expected value of the stochastic variable, $\hat{\theta_i} = \bar{\theta_i}$. The k-NN estimate, on the other hand, involves the utilization of a historical weather dataset, which comprises tuples of $(w^g, \mu_{ig}, c_{i0g}, ..., c_{iNg})$. By employing the k-NN algorithm, the agent can leverage the current $w$ vector to identify the expected value from the $w_g$ closest tuples in the weather dataset,
\begin{align}
\hat{d_{i}} = (\textstyle{\sum}_{w_{g} \in N(w)}^{G} \tfrac{1}{||w_{g}-w||} \mu_{ig})/(\textstyle{\sum}_{w_{g} \in N(w)}^{G} \tfrac{1}{||w_{g}-w||}), \\
\hat{c}_{ij} = (\textstyle{\sum}_{w_{g} \in N(w)}^{G} \tfrac{1}{||w_{g}-w||} c_{ijg})/(\textstyle{\sum}_{w_{g} \in N(w)}^{G} \tfrac{1}{||w_{g}-w||}).
\end{align}

{\bf Customer Positions.} In this study, we introduce two distinct settings for customer positioning: fixed and flexible. In the fixed positioning setting, the positions of customers are predetermined and remain unchanged during the training and inference phases. As a result, the constant component of travel costs, denoted as $\bar{c}_{ij}$, remains constant throughout. This setting may yield improved performance due to the relatively less challenging nature of the problem. It is particularly relevant for businesses operating with fixed customers (e.g. B2B delivery). In contrast, the flexible customer positioning setting allows for varying customer positions in each problem instance. This results in different $\bar{c}_{ij}$ values for each instance (relevant to B2C delivery).

{\bf Delivery Types.} There are two distinct delivery modes of goods: full delivery and partial delivery. In the full delivery mode, the vehicle is only capable of delivering goods to a customer in their entirety, meaning that the goods are indivisible. If a customer has a demand of $\mu_{is}$, the vehicle must deliver the entire quantity at once. If the vehicle's capacity is insufficient, it must perform a recourse action. On the other hand, in the partial delivery mode, the goods can be divided, allowing the vehicle to deliver them in partial quantities.

\section{Architecture Details}

\begin{figure}[!ht]
\centering
\includegraphics[width=0.4\linewidth]{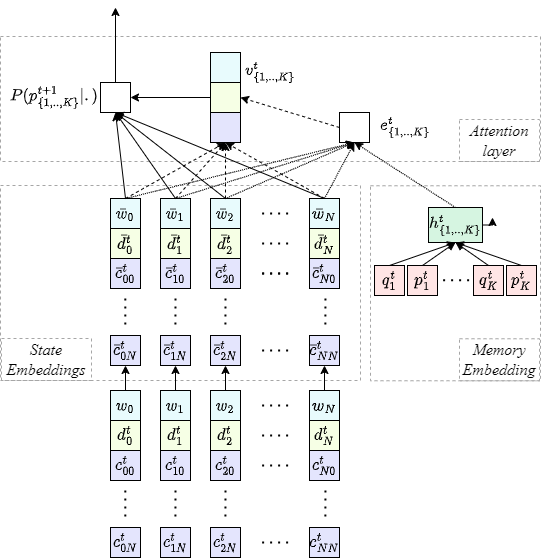}
\caption{\textbf{Network architecture.} The lower section illustrates the model's input, which is composed of three elements: weather data $w$, the varying demand $d_i^t$, and travel costs $c_{ij}^t$. The model generates State Embeddings for each customer and encodes the vehicle's load ($q^t_k$) and current position ($p^t_k$) as the Memory Embedding ($h^t_k$). The embeddings are subsequently combined using an Attention Layer, resulting in probabilities assigned to the nodes. These probabilities represent the likelihood of each node being the next position for vehicle $k$.}
\vskip -8pt
\label{fig:Network}
\end{figure}

{\bf Customer Information.} We present the architectural framework for the models illustrated in Figure \ref{fig:Network}. To obtain continuous representations of customers, referred to as state embeddings, we utilize a 1D convolutional layer with $D$ filters. This layer plays a crucial role in extracting meaningful and continuous representations of the customer-related information $(w, d^t_i, c_{i0}^{t}, ..., c_{iN}^{t})$.

{\bf Vehicle Information.} The vehicle-related information, including the current position $p^{t}_{k}$ and load $q^{t}_{k}$, is processed using an LSTM (Long Short-Term Memory) module with a state size of $D$. This LSTM module captures the temporal dependencies and generates the memory embedding $h^{t}_k$. The memory layer allows the model to retain and utilize information from previous time steps $[1, ..., t-1]$ regarding each vehicle $k$.

{\bf Next Node Probabilities.} To combine the state and memory embeddings effectively, we employ an attention layer. This attention mechanism enables the model to assign importance weights to all node embeddings based on their relevance. The attention layer produces the output $P(p^{t+1}_k|\cdot)$, which represents the probability of each node $i$ (including the depot) being the next node in the route of vehicle $k$.

{\bf Masking.} To ensure the generation of feasible solutions, we employ a masking procedure. Specifically, we assign a value of $-\infty$ to the probability of infeasible position $P(p^{t+1}_{k}|\cdot)$ corresponding to customer nodes whose demand has already been fulfilled. By applying this masking technique, we effectively exclude these already serviced customer nodes from the set of potential next nodes in the route for each vehicle.

\section{Experiments}


{\bf Nodes.} The model was evaluated on four distinct problem sizes involving 10, 20, 50, and 100 customers. To set up the problem instances, we randomly generated the positions of these customers on a 2D map defined by the interval $[0,1] \times [0,1]$. The depot was located at coordinates $(0.5, 0.5)$ as a fixed reference point.

{\bf Variables.} The weather variables $W$ consist of three variables, each following a random uniform distribution within the range of $[-1, 1]$. The noise term $\epsilon$ is generated from a multivariate normal distribution $\mathcal{N}(0, \Sigma)$, where $\Sigma$ is a non-zero covariance matrix. In the experiments, we vary the values of $A_i$, $B_i$, $\Gamma_i$, and $\Phi$. To match these values, we generate the constant and noise components of the stochastic variables accordingly. By employing this generation process, we construct a weather dataset comprising 10,000 scenarios, represented by tuples $(w, \mu_{is}, c_{i0s}, ..., c_{iNs})$ for all $i \in N$. Additionally, we create a separate test dataset consisting of 1,000 scenarios for evaluation purposes.

\subsection{Results}

\begin{table}
\caption{The performance of the ACO algorithm on the SVRP validation dataset. The number of customers are represented in the columns. The results are presented as the average travel cost over the validation dataset, with lower values indicating better performance.}
\label{hyperparam_search_table}
\centering
\begin{small}
\begin{sc}
\scalebox{0.8}{
\begin{tabular}{lrrrr}
\toprule
    ACO parameters &    10 &    20 &     50 &    100 \\
\midrule
         3 / 4 / 20 & 3.73 & 7.61 & 15.88 & 34.86 \\ 
         6 / 8 / 24 & 3.70 & 7.56 & 15.77 & 34.63 \\ 
         4 / 9 / 15 & 3.67 & 7.49 & 15.62 & 34.31 \\ 
         3 / 10/ 12 & \textbf{3.63} & \textbf{7.42} & \textbf{15.46} & \textbf{33.96} \\ 
\bottomrule
\end{tabular}
}
\end{sc}
\end{small}

\caption{The results of the Baseline and RL models on the SVRP test dataset.}

\label{baselines_table}
\centering
\begin{small}
\begin{sc}
\scalebox{0.8}{
\begin{tabular}{lrrrrr}
\\
\toprule
    Baselines &    10 &    20 &     50 &    100 \\
\midrule
         Clarke-Wright & 3.88 & 7.93 & 16.53 & 36.30 \\ 
         Tabu Search & 3.76 & 7.67 & 16.00 & 35.13 \\ 
         Ant-Colony Optimization & 3.63 & 7.42 & 15.46 & 33.96 \\ 
         RL-beam search & \textbf{3.51} & \textbf{7.16} & \textbf{14.93} & \textbf{32.78} \\ 
         
\bottomrule
\end{tabular}
}
\end{sc}
\end{small}
\vskip -8pt
\end{table}

{\bf Baselines.} To establish baseline models, we implemented stochastic approaches based on the works by \cite{Pichpibul2013AHA}, \cite{Li2020AnIT}, \cite{Goel2019VehicleRP}. Since ACO is state-of-the-art algorithm for the SVRP, we conducted a hyperparameter search on validation dataset to identify its optimal configuration for our specific environment. Table~\ref{hyperparam_search_table} displays the results obtained from the search, which involved exploring the pheromone importance, heuristic importance, and the number of ants. Table \ref{baselines_table} illustrates results on test data achieved by baselines and RL with beam search inference ($n_b=3$). The lower values of travel cost show better performance of the model. For fair comparison, the baselines and RL are tested in the same environment configuration (a priori, fixed customers, partial delivery, same distributions and fill rate). RL shows better performance on all sizes with 3.31\%, 3.50\%, 3.43\% and 3.47\% (3.43\% on average) less travel cost compared to ACO results.

{\bf Environment Settings.} To evaluate the sensitivity of the proposed model, we conducted experiments using the default configuration. This configuration included the use of beam search with a beam width of $n_b = 3$ for inference, k-NN variable estimates, fixed customer positions, partial delivery, signal ratios $A, B, \Gamma = 0.6, 0.2, 0.2$ respectively, a fill rate of $\Phi = 0.5$ and a single vehicle $K=1$. The experimental results presented in Table \ref{demand_table} provide insights into the performance of different \textbf{inference strategies} with parameter settings of $n_s = 16$ and $n_b = 3$. The findings indicate that beam search exhibits superior performance compared to the other strategies, highlighting its effectiveness in optimizing the model's output. The comparison presented in Table \ref{demand_table} demonstrates that the k-NN estimation method surpasses the performance of constant demands.

Table \ref{source_table} presents the results obtained from experiments conducted to evaluate the impact of different types of stochasticity. The first experiment involved a fully deterministic model, which naturally demonstrated the best performance. In the second experiment, stochastic demand was introduced, with a ratio of $A, B, \Gamma = 0.6, 0.2, 0.2$. The third experiment introduced stochastic travel cost using the same ratio. The results indicate that stochastic travel cost has a greater impact on performance compared to stochastic demand. This can be attributed to the fact that travel cost is influenced by a larger number of variables compared to demand. In the fourth experiment, both sources of stochasticity were combined, resulting in a more challenging problem setting.

\begin{table}[!ht]
\caption{The performance of inference strategies on the SVRP test dataset. The reported results pertain to both the a priori (Apr.) and reoptimization (Reopt.) settings.}
\label{select_strats_table}
\centering
\begin{small}
\begin{sc}
\scalebox{0.8}{
\begin{tabular}{l cc cc cc cc}
\toprule

 & \multicolumn{2}{c}{{{10}}}
 & \multicolumn{2}{c}{{{20}}}
 & \multicolumn{2}{c}{{{50}}}
 & \multicolumn{2}{c}{{{100}}}\\

 \cmidrule[0.4pt](lr{0.125em}){2-3}%
 \cmidrule[0.4pt](lr{0.125em}){4-5}%
 \cmidrule[0.4pt](lr{0.125em}){6-7}%
 \cmidrule[0.4pt](lr{0.125em}){8-9}%
 
 Inference strategies
 & Apr.
 & Reopt.
 & Apr.
 & Reopt.
 & Apr.
 & Reopt.
 & Apr.
 & Reopt.\\
 
\midrule
     RL-greedy & 3.98 & 3.86 & 8.12 & 7.88 & 16.93 & 16.42 & 37.18 & 36.06 \\ 
     RL-sampling & 3.69 & 3.58 & 7.54 & 7.32 & 15.73 & 15.26 & 34.54 & 33.50 \\ 
     RL-beam-search  & \textbf{3.51} & \textbf{3.40} & \textbf{7.16} & \textbf{6.95} & \textbf{14.93} & \textbf{14.48} & \textbf{32.78} & \textbf{31.80} \\ 
\bottomrule
\end{tabular}
}
\end{sc}
\end{small}
\caption{The results obtained from two different types of variable estimates on the SVRP test dataset.}
\centering
\label{demand_table}

    \scalebox{0.8}{
        \begin{tabular}{l cc cc cc cc} 
\toprule

 & \multicolumn{2}{c}{{{10}}}
 & \multicolumn{2}{c}{{{20}}}
 & \multicolumn{2}{c}{{{50}}}
 & \multicolumn{2}{c}{{{100}}}\\

 \cmidrule[0.4pt](lr{0.125em}){2-3}%
 \cmidrule[0.4pt](lr{0.125em}){4-5}%
 \cmidrule[0.4pt](lr{0.125em}){6-7}%
 \cmidrule[0.4pt](lr{0.125em}){8-9}%
 
 Variable Estimate 
 & Apr.
 & Reopt.
 & Apr.
 & Reopt.
 & Apr.
 & Reopt.
 & Apr.
 & Reopt.\\
 
\midrule
          Constant variable estimate & 3.76 & 3.64 & 7.67 & 7.44 & 16.0 & 15.52 & 35.13 & 34.07 \\ 
          kNN variable estimate & \textbf{3.51} & \textbf{3.40} & \textbf{7.16} & \textbf{6.95} & \textbf{14.93} & \textbf{14.48} & \textbf{32.78} & \textbf{31.80} \\ 
 
        \end{tabular}
    }

\caption{The results of various stochasticity sources on the SVRP test dataset.}
\centering
\label{source_table}
    \scalebox{0.8}{
        \begin{tabular}{l cc cc cc cc} 
\\
\toprule

 & \multicolumn{2}{c}{{{10}}}
 & \multicolumn{2}{c}{{{20}}}
 & \multicolumn{2}{c}{{{50}}}
 & \multicolumn{2}{c}{{{100}}}\\

 \cmidrule[0.4pt](lr{0.125em}){2-3}%
 \cmidrule[0.4pt](lr{0.125em}){4-5}%
 \cmidrule[0.4pt](lr{0.125em}){6-7}%
 \cmidrule[0.4pt](lr{0.125em}){8-9}%
 
 Stochasticity Sources
 & Apr.
 & Reopt.
 & Apr.
 & Reopt.
 & Apr.
 & Reopt.
 & Apr.
 & Reopt.\\
 
\midrule
     No Stochasticity & 2.88 & 2.79 & 5.88 & 5.71 & 12.26 & 11.90 & 26.93 & 26.12 \\ 
     Stochastic Demand & 3.13 & 3.04 & 6.39 & 6.20 & 13.33 & 12.93 & 29.27 & 28.39 \\ 
     Stochastic Travel Cost & 3.38 & 3.28 & 6.90 & 6.70 & 14.40 & 13.96 & 31.61 & 30.67 \\ 
     Stochastic Demand / Travel Cost & 3.51 & 3.40 & 7.16 & 6.95 & 14.93 & 14.48 & 32.78 & 31.80 \\ 

 \bottomrule
        \end{tabular}
    }
\vskip -8pt
\end{table}

Table \ref{signal_ratio_table} presents the impact of varying signal ratio settings on the performance of the model. The findings indicate that increasing the signal from the weather variable leads to a reduction in travel costs. This suggests that the RL model is capable of capturing and leveraging the implicit influence of weather on demands and travel times. It is worth noting that this impact may vary across different customers, yet the model can effectively utilize this information to enhance routing strategies. The performance of the model was evaluated through experiments examining the influence of \textbf{customer positions} on the outcomes, as depicted in Table \ref{position}. The findings suggest that fixed customer positions yield superior performance on the test data. 

\begin{table}[!ht]
\centering
\caption{The results obtained from various signal ratios on the SVRP test dataset.}
\label{signal_ratio_table}
    \scalebox{0.8}{
        \begin{tabular}{c cc cc cc cc} 
\\
\toprule

 & \multicolumn{2}{c}{{{10}}}
 & \multicolumn{2}{c}{{{20}}}
 & \multicolumn{2}{c}{{{50}}}
 & \multicolumn{2}{c}{{{100}}}\\

 \cmidrule[0.4pt](lr{0.125em}){2-3}%
 \cmidrule[0.4pt](lr{0.125em}){4-5}%
 \cmidrule[0.4pt](lr{0.125em}){6-7}%
 \cmidrule[0.4pt](lr{0.125em}){8-9}%
 
 $\mathrm{A}/\mathrm{B}/\Gamma$ 
 & Apr.
 & Reopt.
 & Apr.
 & Reopt.
 & Apr.
 & Reopt.
 & Apr.
 & Reopt.\\
 
\midrule
         0.8 / 0.0 / 0.2  & 3.41 & 3.31 & 6.97 & 6.76 & 14.53 & 14.09 & 31.91 & 30.95 \\ 
         0.8 / 0.2 / 0.0 & \textbf{3.35} & \textbf{3.25} & \textbf{6.84} & \textbf{6.64} & \textbf{14.26} & \textbf{13.84} & \textbf{31.32} & \textbf{30.38} \\ 
         0.6 / 0.2 / 0.2 & 3.51 & 3.40 & 7.16 & 6.95 & 14.93 & 14.48 & 32.78 & 31.80 \\ 
         0.4 / 0.3 / 0.3 & 3.76 & 3.64 & 7.67 & 7.44 & 16.00 & 15.52 & 35.13 & 34.07 \\ 
        \end{tabular}
    }
\centering
\caption{The results obtained from two types of positions on the SVRP test dataset.}
\label{position}
    \scalebox{0.8}{
        \begin{tabular}{l cc cc cc cc} 
\\
\toprule

 & \multicolumn{2}{c}{{{10}}}
 & \multicolumn{2}{c}{{{20}}}
 & \multicolumn{2}{c}{{{50}}}
 & \multicolumn{2}{c}{{{100}}}\\

 \cmidrule[0.4pt](lr{0.125em}){2-3}%
 \cmidrule[0.4pt](lr{0.125em}){4-5}%
 \cmidrule[0.4pt](lr{0.125em}){6-7}%
 \cmidrule[0.4pt](lr{0.125em}){8-9}%

  Customers Position
 & Apr.
 & Reopt.
 & Apr.
 & Reopt.
 & Apr.
 & Reopt.
 & Apr.
 & Reopt.\\
 
\midrule
         Flexible customer positions & 3.88 & 3.77 & 7.93 & 7.69 & 16.53 & 16.03 & 36.30 & 35.21 \\ 
         Fixed customer positions & \textbf{3.51} & \textbf{3.40} & \textbf{7.16} & \textbf{6.95} & \textbf{14.93} & \textbf{14.48} & \textbf{32.78} & \textbf{31.80} \\ 
        \end{tabular}
    }
\centering
\caption{The results obtained from various fill rates on the SVRP test dataset.}
\label{fill_rate_table}
    \scalebox{0.65}{
        \begin{tabular}{c cc cc cc cc} 
\\
\toprule

 & \multicolumn{2}{c}{{{10}}}
 & \multicolumn{2}{c}{{{20}}}
 & \multicolumn{2}{c}{{{50}}}
 & \multicolumn{2}{c}{{{100}}}\\

 \cmidrule[0.4pt](lr{0.125em}){2-3}%
 \cmidrule[0.4pt](lr{0.125em}){4-5}%
 \cmidrule[0.4pt](lr{0.125em}){6-7}%
 \cmidrule[0.4pt](lr{0.125em}){8-9}%
 
 Fill Rate 
 & Apr.
 & Reopt.
 & Apr.
 & Reopt.
 & Apr.
 & Reopt.
 & Apr.
 & Reopt.\\
 
\midrule
         0.1 & 3.88 & 3.77 & 7.93 & 7.69 & 16.53 & 16.03 & 36.30 & 35.21 \\ 
         0.5 & 3.51 & 3.40 & 7.16 & 6.95 & 14.93 & 14.48 & 32.78 & 31.80 \\ 
         0.9 & \textbf{3.29} & \textbf{3.19} & \textbf{6.71} & \textbf{6.51} & \textbf{14.00} & \textbf{13.58} & \textbf{30.74} & \textbf{29.81} \\ 
         
        \end{tabular}
    }
\vskip -8pt
\end{table}

We tested the performance variation between baseline models and RL models under correlated and uncorrelated (demand, travel cost) settings (Figure~1 of the Supplementary Material). It is evident from the results that the baseline models lack the ability to capture and utilize the correlations between customers, leading to suboptimal routing decisions. In contrast, the RL models demonstrate the capability to exploit these correlations, resulting in more cost-effective routes.

Table~\ref{fill_rate_table} presents the findings from experiments conducted with different values of the signal ratio. The results indicate that increasing the fill rate, which corresponds to a larger capacity of the vehicle, facilitates the routing task. This is reflected in the improved performance of the model, as it is able to identify more optimal routes under such conditions. The experiments were conducted with partial delivery, which allows the agent to deliver goods in a more convenient manner, leading to a comparatively easier problem setting. As reflected in the results, the performance of the model is higher under the partial delivery setting, indicating its ability to leverage this condition to identify more efficient routes (Table~1 of the Supplementary Material).

{\bf Multi-Vehicle Scenarios.} Table \ref{mutiple_vehicles} presents the results of experiments conducted to assess the impact of the number of vehicles on the performance of the model. Theoretically, increasing the number of vehicles should simplify the problem by enabling the division of a large route into multiple smaller routes, resulting in fewer recourse actions and reduced total length. Consistent with this notion, our findings demonstrate that as the number of vehicles increases, the model's performance improves.

\begin{table}
\caption{The results obtained from various number of vehicles on the SVRP test dataset.}
\centering
\label{mutiple_vehicles}

    \scalebox{0.8}{
        \begin{tabular}{c cc cc cc cc} 
\\
\toprule

 & \multicolumn{2}{c}{{{10}}}
 & \multicolumn{2}{c}{{{20}}}
 & \multicolumn{2}{c}{{{50}}}
 & \multicolumn{2}{c}{{{100}}}\\

 \cmidrule[0.4pt](lr{0.125em}){2-3}%
 \cmidrule[0.4pt](lr{0.125em}){4-5}%
 \cmidrule[0.4pt](lr{0.125em}){6-7}%
 \cmidrule[0.4pt](lr{0.125em}){8-9}%
 
 Vehicle Number
 & Apr.
 & Reopt.
 & Apr.
 & Reopt.
 & Apr.
 & Reopt.
 & Apr.
 & Reopt.\\
 \midrule
     1 & 3.51 & 3.40 & 7.16 & 6.95 & 14.93 & 14.48 & 32.78 & 31.80 \\ 
     2 & 3.43 & 3.33 & 7.01 & 6.80 & 14.61 & 14.17 & 32.08 & 31.12 \\ 
     3 & 3.41 & 3.31 & 6.96 & 6.75 & 14.52 & 14.08 & 31.88 & 30.92 \\ 
     5 & \textbf{3.40} & \textbf{3.29} & \textbf{6.94} & \textbf{6.73} & \textbf{14.46} & \textbf{14.03} & \textbf{31.76} & \textbf{30.81} \\ 
        \end{tabular}
    }

\caption{Inference time results in seconds on the SVRP test dataset.}
\centering
\label{comp_times}
    \scalebox{0.8}{
        \begin{tabular}{l c c c c} 
\\
\toprule
 Baseline
 & 10
 & 20
 & 50
 & 100\\

\midrule
    Clarke-Wright &	0.005 &	0.016 &	0.058 &	0.185 \\ 
    Tabu Search &	0.964 &	2.936 &	8.672 &	57.12 \\ 
    Ant-Colony Optimization &	1.230 &	4.291 &	27.64 &	123.7 \\ 
    RL-greedy &	0.061 &	0.121 &	0.191 &	0.394 \\ 
    RL-sampling &	0.064 &	0.125 &	0.232 &	0.405 \\ 
    RL-beam search &	0.072 &	0.178 &	0.293 &	0.416 \\ 
        \end{tabular}
    }
\vskip -8pt
\end{table}

{\bf Time Complexity.} Table \ref{comp_times} presents the inference times of RL models and baselines for a single problem instance. It is observed that RL and CW models exhibit inference times that are suitable for real-time solving, indicating their efficiency in finding solutions for industry (hundreds of customers). On the other hand, ACO and Tabu search models demonstrate inference times that have an exponential dependence on the problem size, suggesting that their computational requirements increase significantly as the problem becomes larger.

\section{Conclusion}

In conclusion, the Stochastic Vehicle Routing Problem (SVRP) is a challenging yet significant task in the industry. It is characterized by two crucial sources of stochasticity: stochastic customer demands and stochastic travel costs. In this study, we have presented a formulation of the problem that incorporates both sources of uncertainty. Furthermore, we have proposed an RL agent architecture along with a training methodology to effectively address the formulated problem. In real-world industrial applications, companies often leverage available information that can potentially influence stochastic demand and travel costs. Weather information is one such example that can introduce correlation among customers. In our study, we integrated this type of information into the environment and architecture of our model. The results revealed that the RL agent, when provided with access to weather information, was able to achieve \textbf{routing strategies that outperformed those without such information by up to 2\%}.

This finding highlights the significance of incorporating relevant external factors, such as weather, into the modeling and decision-making processes of stochastic routing problems. By utilizing this additional information, the RL agent was able to effectively capture and exploit the correlations among customers, resulting in improved routing strategies and ultimately leading to enhanced operational efficiency in real-world scenarios. We conducted a set of evaluations of the proposed model in various settings relevant to the VRP in industrial applications. The results demonstrate the robustness of the model in learning effective routing strategies across all tested scenarios. Furthermore, we compared the model to a state-of-the-art approach introduced by \cite{Goel2019VehicleRP}, and our\textbf{ RL agent exhibited a superior performance, achieving a significant 3.43\% improvement} in travel cost.

The primary focus of this work was to introduce the first RL model capable of addressing the VRP with stochastic demands and travel costs and to showcase its superiority over classical models commonly employed in industrial settings. The results indicate that RL has the potential to replace traditional approaches and significantly reduce routing costs, consequently leading to reductions in logistics expenses and negative environmental impact. Future research in this field can delve deeper into the application of RL for the SVRP and explore the development of alternative architectures and training methods to further enhance the performance of the model. These \textbf{efforts can contribute to the advancement of efficient and sustainable routing solutions} in logistics and transportation domains.

\newpage
\bibliography{acml23}

\begin{thebibliography}{29}
\providecommand{\natexlab}[1]{#1}
\providecommand{\url}[1]{\texttt{#1}}
\expandafter\ifx\csname urlstyle\endcsname\relax
  \providecommand{\doi}[1]{doi: #1}\else
  \providecommand{\doi}{doi: \begingroup \urlstyle{rm}\Url}\fi

\bibitem[Andrychowicz et~al.(2020)Andrychowicz, Baker, Chociej, J{\'o}zefowicz,
  McGrew, Pachocki, Petron, Plappert, Powell, Ray, Schneider, Sidor, Tobin,
  Welinder, Weng, and Zaremba]{Andrychowicz2020LearningDI}
Marcin Andrychowicz, Bowen Baker, Maciek Chociej, Rafal J{\'o}zefowicz, Bob
  McGrew, Jakub~W. Pachocki, Arthur Petron, Matthias Plappert, Glenn Powell,
  Alex Ray, Jonas Schneider, Szymon Sidor, Joshua Tobin, Peter Welinder, Lilian
  Weng, and Wojciech Zaremba.
\newblock Learning dexterous in-hand manipulation.
\newblock \emph{The International Journal of Robotics Research}, 39:\penalty0
  20 -- 3, 2020.

\bibitem[Bello et~al.(2017)Bello, Pham, Le, Norouzi, and
  Bengio]{Bello2017NeuralCO}
Irwan Bello, Hieu Pham, Quoc~V. Le, Mohammad Norouzi, and Samy Bengio.
\newblock Neural combinatorial optimization with reinforcement learning.
\newblock \emph{ArXiv}, abs/1611.09940, 2017.

\bibitem[Bomboi et~al.(2021)Bomboi, Buchheim, and Pruente]{Bomboi2021OnTS}
Federica Bomboi, Christoph Buchheim, and Jonas Pruente.
\newblock On the stochastic vehicle routing problem with time windows,
  correlated travel times, and time dependency.
\newblock \emph{4OR}, pages 1--23, 2021.

\bibitem[Christiansen and Lysgaard(2007)]{Christiansen2007ABA}
Christian~H. Christiansen and Jens Lysgaard.
\newblock A branch-and-price algorithm for the capacitated vehicle routing
  problem with stochastic demands.
\newblock \emph{Oper. Res. Lett.}, 35:\penalty0 773--781, 2007.

\bibitem[Cordeau et~al.(2007)Cordeau, Laporte, Savelsbergh, and Vigo]{inbook}
Jean-Fran{\c{c}}ois Cordeau, Gilbert Laporte, Martin~WP Savelsbergh, and
  Daniele Vigo.
\newblock Vehicle routing.
\newblock \emph{Handbooks in operations research and management science},
  14:\penalty0 367--428, 2007.

\bibitem[Dantzig and Ramser(1959)]{Dantzig1959TheTD}
George~B. Dantzig and John~Hubert Ramser.
\newblock The truck dispatching problem.
\newblock \emph{Management Science}, 6:\penalty0 80--91, 1959.

\bibitem[Gauvin et~al.(2014)Gauvin, Desaulniers, and Gendreau]{Gauvin2014ABA}
Charles Gauvin, Guy Desaulniers, and Michel Gendreau.
\newblock A branch-cut-and-price algorithm for the vehicle routing problem with
  stochastic demands.
\newblock \emph{Comput. Oper. Res.}, 50:\penalty0 141--153, 2014.

\bibitem[Gendreau et~al.(1996)Gendreau, Laporte, and S{\'e}guin]{GENDREAU19963}
Michel Gendreau, Gilbert Laporte, and Ren{\'e} S{\'e}guin.
\newblock Stochastic vehicle routing.
\newblock \emph{European Journal of Operational Research}, 88\penalty0
  (1):\penalty0 3--12, 1996.

\bibitem[Goel et~al.(2019)Goel, Maini, and Bansal]{Goel2019VehicleRP}
Rajeev~Kumar Goel, Raman Maini, and Sandhya Bansal.
\newblock Vehicle routing problem with time windows having stochastic customers
  demands and stochastic service times: Modelling and solution.
\newblock \emph{J. Comput. Sci.}, 34:\penalty0 1--10, 2019.

\bibitem[Helsgaun(2017)]{Helsgaun2017AnEO}
Keld Helsgaun.
\newblock An extension of the lin-kernighan-helsgaun tsp solver for constrained
  traveling salesman and vehicle routing problems: Technical report.
\newblock 2017.

\bibitem[Iklassov et~al.(2023)Iklassov, Sobirov, Solozabal, and
  Tak{\'{a}}c]{iklassovSST23}
Zangir Iklassov, Ikboljon Sobirov, Ruben Solozabal, and Martin Tak{\'{a}}c.
\newblock Reinforcement learning approach to stochastic vehicle routing problem
  with correlated demands.
\newblock \emph{{IEEE} Access}, 11:\penalty0 87958--87969, 2023.
\newblock \doi{10.1109/ACCESS.2023.3306076}.
\newblock URL \url{https://doi.org/10.1109/ACCESS.2023.3306076}.

\bibitem[Joe and Lau(2020)]{Joe2020DeepRL}
Waldy Joe and Hoong~Chuin Lau.
\newblock Deep reinforcement learning approach to solve dynamic vehicle routing
  problem with stochastic customers.
\newblock In \emph{ICAPS}, 2020.

\bibitem[Laporte(2009)]{Laporte2009FiftyYO}
Gilbert Laporte.
\newblock Fifty years of vehicle routing.
\newblock \emph{Transp. Sci.}, 43:\penalty0 408--416, 2009.

\bibitem[Laporte and Louveaux(1993)]{Laporte1993TheIL}
Gilbert Laporte and François~V. Louveaux.
\newblock The integer l-shaped method for stochastic integer programs with
  complete recourse.
\newblock \emph{Oper. Res. Lett.}, 13:\penalty0 133--142, 1993.

\bibitem[Li and Li(2020)]{Li2020AnIT}
Guoming Li and Junhua Li.
\newblock An improved tabu search algorithm for the stochastic vehicle routing
  problem with soft time windows.
\newblock \emph{IEEE Access}, 8:\penalty0 158115--158124, 2020.

\bibitem[Louveaux(1998)]{Louveaux1998AnIT}
François~V. Louveaux.
\newblock An introduction to stochastic transportation models.
\newblock 1998.

\bibitem[Lu et~al.(2020)Lu, Zhang, and Yang]{Lu2020ALI}
Hao Lu, Xingwen Zhang, and Shuang Yang.
\newblock A learning-based iterative method for solving vehicle routing
  problems.
\newblock In \emph{ICLR}, 2020.

\bibitem[Mustakhov et~al.(2023)Mustakhov, Akhmetbek, and
  Bogyrbayeva]{Mustakhov2023DeepRL}
Taukekhan Mustakhov, Yernar Akhmetbek, and Aigerim Bogyrbayeva.
\newblock Deep reinforcement learning for stochastic dynamic vehicle routing
  problem.
\newblock \emph{2023 17th International Conference on Electronics Computer and
  Computation (ICECCO)}, pages 1--5, 2023.
\newblock URL \url{https://api.semanticscholar.org/CorpusID:259158985}.

\bibitem[Nazari et~al.(2018)Nazari, Oroojlooy, Snyder, and
  Tak{\'a}{\v{c}}]{nazari2018reinforcement}
Mohammadreza Nazari, Afshin Oroojlooy, Lawrence~V Snyder, and Martin
  Tak{\'a}{\v{c}}.
\newblock Reinforcement learning for solving the vehicle routing problem.
\newblock In \emph{Conference on Neural Information Processing Systems, NeurIPS
  2018}, 2018.

\bibitem[Niu et~al.(2021)Niu, Kong, Wen, Cao, and hua Xiao]{Niu2021AnIL}
Yunyun Niu, Detian Kong, Rong Wen, Zhiguang Cao, and Jian hua Xiao.
\newblock An improved learnable evolution model for solving multi-objective
  vehicle routing problem with stochastic demand.
\newblock \emph{Knowl. Based Syst.}, 230:\penalty0 107378, 2021.

\bibitem[Niu et~al.(2022)Niu, Shao, hua Xiao, Song, and
  Cao]{Niu2022MultiobjectiveEA}
Yunyun Niu, Jie Shao, Jian hua Xiao, Wen Song, and Zhiguang Cao.
\newblock Multi-objective evolutionary algorithm based on rbf network for
  solving the stochastic vehicle routing problem.
\newblock \emph{Inf. Sci.}, 609:\penalty0 387--410, 2022.

\bibitem[Oroojlooyjadid et~al.(2022)Oroojlooyjadid, Nazari, Snyder, and
  Tak{\'a}c]{Oroojlooyjadid2022ADQ}
Afshin Oroojlooyjadid, M.~Nazari, Lawrence~V. Snyder, and Martin Tak{\'a}c.
\newblock A deep q-network for the beer game: Deep reinforcement learning for
  inventory optimization.
\newblock \emph{Manuf. Serv. Oper. Manag.}, 24:\penalty0 285--304, 2022.

\bibitem[Pichpibul and Kawtummachai(2013)]{Pichpibul2013AHA}
Tantikorn Pichpibul and Ruengsak Kawtummachai.
\newblock A heuristic approach based on clarke-wright algorithm for open
  vehicle routing problem.
\newblock \emph{The Scientific World Journal}, 2013, 2013.

\bibitem[Rajabi-Bahaabadi et~al.(2021)Rajabi-Bahaabadi, Shariat, Babaei, and
  Vigo]{Bahaabadi2021}
Mojtaba Rajabi-Bahaabadi, Afshin Shariat, Mohsen Babaei, and Daniele Vigo.
\newblock Reliable vehicle routing problem in stochastic networks with
  correlated travel times.
\newblock \emph{Operational Research}, 03 2021.
\newblock \doi{10.1007/s12351-019-00452-w}.

\bibitem[Secomandi(2000)]{Secomandi2000ComparingNP}
Nicola Secomandi.
\newblock Comparing neuro-dynamic programming algorithms for the vehicle
  routing problem with stochastic demands.
\newblock \emph{Comput. Oper. Res.}, 27:\penalty0 1201--1225, 2000.

\bibitem[Silver et~al.(2017)Silver, Schrittwieser, Simonyan, Antonoglou, Huang,
  Guez, Hubert, Baker, Lai, Bolton, et~al.]{silver2017mastering}
David Silver, Julian Schrittwieser, Karen Simonyan, Ioannis Antonoglou, Aja
  Huang, Arthur Guez, Thomas Hubert, Lucas Baker, Matthew Lai, Adrian Bolton,
  et~al.
\newblock Mastering the game of go without human knowledge.
\newblock \emph{nature}, 550\penalty0 (7676):\penalty0 354--359, 2017.

\bibitem[Toth and Vigo(2014)]{Toth2014VehicleRP}
Paolo Toth and Daniele Vigo.
\newblock Vehicle routing: Problems, methods, and applications, second edition.
\newblock 2014.

\bibitem[Tricks(2018)]{Tricks2018AND}
Learns~Old Tricks.
\newblock A new dog learns old tricks : Rl finds classic optimization
  algorithms.
\newblock 2018.

\bibitem[Wang(2021)]{Wang2021AlphaTLT}
Q.~Wang.
\newblock Alpha-t: Learning to traverse over graphs with an alphazero-inspired
  self-play framework.
\newblock 2021.

\end{thebibliography}
\appendix

\end{document}